\documentclass[10pt,twocolumn,letterpaper]{article}

\usepackage{cvpr}              
\usepackage{graphicx}
\definecolor{cvprblue}{rgb}{0.21,0.49,0.74}
\usepackage[pagebackref,breaklinks,colorlinks,allcolors=cvprblue]{hyperref}

\def\confName{CVPR}
\def\confYear{2026}

\title{TorchTraceAP: A New Benchmark Dataset for Detecting Performance Anti-Patterns in Computer Vision Models}

\author{Hanning Chen$^{1}$\thanks{This work was done during a Meta internship.} \quad Keyu Man$^{2}$ \quad Kevin Zhu$^{2}$ \quad Chenguang Zhu$^{2}$ \quad Haonan Li$^{3}$ \\ 
\quad Tongbo Luo$^{2}$ \quad Xizhou Feng$^{2}$ \quad Wei Sun$^{2}$ \quad Sreen Tallam$^{2}$ \quad Mohsen Imani$^{1}$ \quad Partha Kanuparthy$^{2}$ 
\vspace{0.3em} 
\\
{\normalsize $^1$ University of California, Irvine, CA, USA} \\
{\normalsize $^2$ Meta, Menlo Park, CA, USA} \\
{\normalsize $^3$ University of California, Riverside, CA, USA} \\
{\tt\small \{hanningc\}@uci.edu}
}

\begin{document}
\maketitle

\begin{abstract}
Identifying and addressing performance anti-patterns in machine learning (ML) models is critical for efficient training and inference, but it typically demands deep expertise spanning system infrastructure, ML models and kernel development. While large tech companies rely on dedicated ML infrastructure engineers to analyze torch traces and benchmarks, such resource-intensive workflows are largely inaccessible to computer vision researchers in general. Among the challenges, pinpointing problematic trace segments within lengthy execution traces remains the most time-consuming task, and is difficult to automate with current ML models, including LLMs. In this work, we present the first benchmark dataset specifically designed to evaluate and improve ML models' ability to detect anti-patterns in traces. Our dataset contains over 600 PyTorch traces from diverse computer vision models—classification, detection, segmentation, and generation—collected across multiple hardware platforms. We also propose a novel iterative approach: a lightweight ML model first detects trace segments with anti-patterns, followed by a large language model (LLM) for fine-grained classification and targeted feedback. Experimental results demonstrate that our method significantly outperforms unsupervised clustering and rule-based statistical techniques for detecting anti-pattern regions. Our method also effectively compensates LLM's limited context length and reasoning inefficiencies.

\end{abstract}

\section{Introduction}
\begin{figure}[t]
    \centering
    \includegraphics[width=\linewidth]{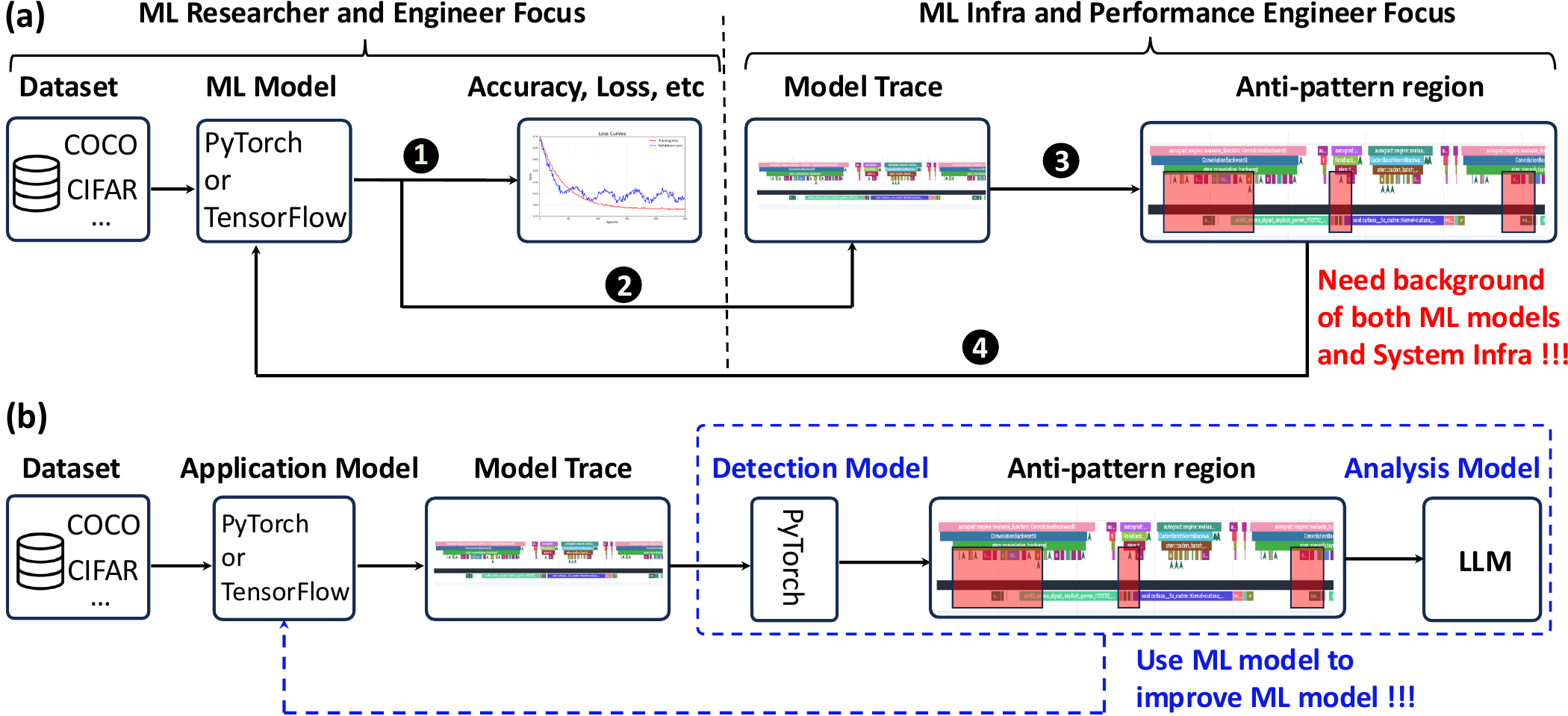}
    \caption{(a) Traditionally ML model development involves both machine learning engineer and system infra engineer. (b) This work we propose use ML model to improve ML model's performance.}
    \label{fig:motivation}
\end{figure}

Detecting inefficiencies in the training and inference of computer vision (CV) models is a critical task for deploying these models in real-world applications~\cite{liang2024fine}. On identical hardware platforms, a well-optimized model pipeline can achieve up to an $8\times$ in both training and inference compared with the unoptimized version~\cite{torch_hta}. For large or computationally intensive models, such a significant performance improvement can transform an otherwise unusable application into a practical and deployable solution.

Despite its importance, torch trace analysis and anti-pattern detection remain challenging for most computer vision researchers. The typical workflow uses the PyTorch profiler~\cite{paszke2019pytorch} to collect torch traces, visualizes them with tools like TensorBoard or Perfetto~\cite{google_perfetto}, and identifies anomalous segments (Figure~\ref{fig:motivation}(a)). This is difficult for two reasons: (1) interpreting torch traces requires expertise in ML, computer architecture, and system profiling—knowledge often beyond the scope of CV researchers; and (2) the process is time-consuming, as a single profiling run can generate thousands of CPU and GPU events (e.g., ~8,000 for ResNet-50~\cite{he2016deep}), with even greater complexity for advanced models like vision transformers~\cite{dosovitskiy2020image} used in segmentation~\cite{kirillov2023segment}.

As shown in Figure~\ref{fig:motivation}(a), a common industry practice is to divide responsibilities: machine learning engineers and researchers focus on model accuracy and application development, while ML infrastructure engineers concentrate on profiling and optimizing model efficiency. However, for many academic and smaller research groups, it is not feasible to form such multidisciplinary teams. As a result, many CV models are developed with a primary focus on novel applications and accuracy improvements, often at the expense of the efficiency and optimization of the system.

Recent advances in large language models (LLMs)~\cite{chang2024survey}, retrieval augmentation (RAG)~\cite{arslan2024survey}, and LLM agents~\cite{huang2024understanding} have enabled applications in code generation~\cite{liu2024exploring}, program analysis~\cite{yang2024enhancing, li2024enhancing}, and performance debugging~\cite{lee2024unified}. Building on these developments and as shown in Figure~\ref{fig:motivation}(b), we propose leveraging ML models to improve other ML models by simplifying torch trace anti-pattern detection, making performance optimization accessible to smaller research groups. However, LLMs’ limited context length and slow inference make direct analysis of large torch traces impractical. Inspired by anomaly detection in videos~\cite{sultani2018real} and time-series data~\cite{lee2021weakly}, we advocate a two-stage approach: a lightweight ML model first identifies suspicious trace segments, which are then analyzed by an LLM. \textbf{To our knowledge, no existing datasets or research address training anti-pattern detectors specifically for torch traces.}

To address this gap, we introduce \textbf{TorchTraceAP}, a benchmark dataset for detecting performance anti-patterns in computer vision models. Our main contributions are:

\begin{itemize}
    \item TorchTraceAP is the first dataset and benchmark for ML-based detection of anti-patterns from torch traces across diverse CV models.
    \item We develop a customized anomaly detection model with novel event and window encoders, and an anomaly predictor tailored for torch trace data.
    \item We propose an interactive, feedback-driven system that combines a lightweight ML model for anomaly detection with an LLM for detailed analysis.
    \item Our approach outperforms rule-based methods by 15\% in detecting Torch anti-patterns, and the LLM achieves 75\% accuracy in anti-pattern type classification.
\end{itemize}


\begin{figure*}[t]
    \centering
    \includegraphics[width=0.86\linewidth]{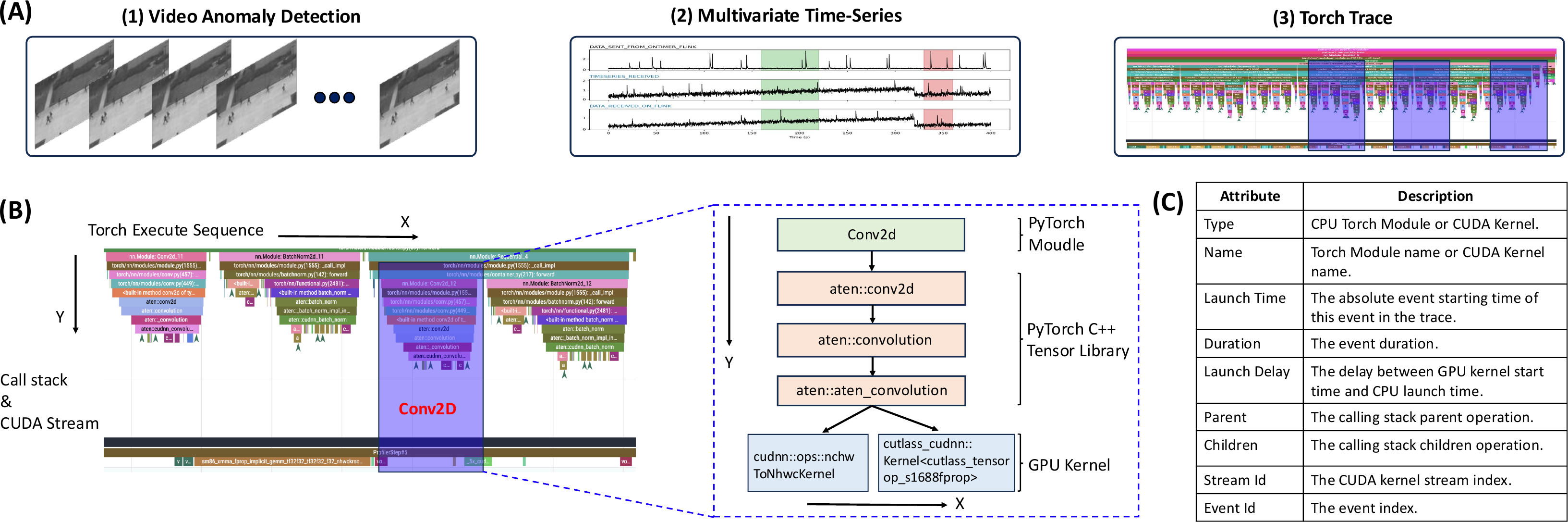}
    \caption{(A) Comparison between video anomaly detection~\cite{sultani2018real}, multivariate time-series anomaly detection~\cite{lee2021weakly}, and torch trace anti-patterns detection (this work). (B) Torch trace example and illustration of torch model calling stack. (C) Torch trace event attributes.}
    \label{fig:background}
\end{figure*}

\section{Challenges \& Related Work}

In the current machine learning and computer vision communities, the tasks and datasets most closely related to TorchTraceAP are \textbf{video anomaly detection} (VAD) and \textbf{multivariate time series anomaly detection} (M-TSAD). Figure~\ref{fig:background}(A) presents a visual comparison between video, multivariate time series signals, and torch traces. Previous works~\cite{abdalla2025video, zamanzadeh2024deep, wu2024deep} have explored semi-supervised~\cite{georgescu2021background,liu2023generating}, weakly-supervised~\cite{tian2021weakly, lee2021weakly}, and unsupervised methods~\cite{del2016discriminative, yu2022deep, xu2018unsupervised} for detecting anomalous image or signal segments at various levels, including video and time-series~\cite{zanella2024delving}, image or signal segment~\cite{tian2021weakly}, as well as per-window and point-level~\cite{su2019robust} detection. Generally, the more fine-grained the anomaly detection task, the more training information is required.

While torch traces display time-series patterns, detecting anti-patterns (anomalies) is notably more difficult. This is due to two main challenges: (1) Unlike VAD and M-TSAD, which have established feature extractors for video and time-series data (e.g., C3D RGB~\cite{tran2015learning}, I3D RGB~\cite{carreira2017quo}), there is currently no standard feature extractor for torch traces. (2) Many VAD and M-TSAD methods use multiple-instance learning (MIL)~\cite{sultani2018real}, assuming positive samples lack anomalies. This assumption does not hold for torch traces, as even normal models may contain inefficient kernel patterns unless manually optimized, complicating loss design and dataset preparation.

To address these challenges, we introduce TorchTraceAP, a benchmark for developing and evaluating anti-pattern detection in torch traces (Figure~\ref{fig:motivation}.(b)). We also propose an iterative detection framework with a custom feature encoder, tailored loss function, and LLM backend. Section~\ref{sec:trace_background} and Section~\ref{sec:anti_pattern} provide background, Section~\ref{sec:trace_dataset} describes the dataset, and Section~\ref{sec:method} details our framework.

\section{Background}
\subsection{Torch Trace} \label{sec:trace_background}

Figure~\ref{fig:background}.(B) provides a visual snapshot of a ResNet18 torch trace~\cite{paszke2019pytorch}. In essence, a torch trace is a two-dimensional representation of the sequence and hierarchy of events(operations) that occur during the execution of a PyTorch model.
\begin{itemize}
    \item \textbf{X-axis (Horizontal):} Represents the chronological order in which events (operations) are executed. For example, as ResNet18 processes an input, its layers are executed one after another, and each of these executions is recorded as an event along the X-axis.
    \item \textbf{Y-axis (Vertical):} Represents the call stack of operations. This shows how high-level operations are broken down into lower-level function calls. For instance, when PyTorch executes a \texttt{Conv2d} (2D convolution) operation, it internally calls lower-level CPU APIs such as \texttt{aten::conv2d}, which may further call \texttt{aten::convolution}, and so on. If the operation is run on a GPU, these CPU calls will also trigger the launch of corresponding CUDA kernels.
\end{itemize}
Thus, a torch trace can be thought of as a collection of CPU and GPU events, organized in a two-dimensional space: time (X-axis) and call hierarchy (Y-axis).
Figure~\ref{fig:background}(C) further illustrates the meta-attributes associated with each torch trace event. Each event contains several key attributes, such as:
\begin{itemize}
    \item \textbf{Type:} The kind of operation (e.g., CPU operation, CUDA kernel).
    \item \textbf{Name:} The specific operation or function being executed.
    \item \textbf{Launch Time:} When the event was initiated.
    \item \textbf{Parent/Children:} Relationships indicating which event triggered (parent) or was triggered by (children) this event.
    \item \textbf{Stream Index (for CUDA kernels):} Indicates which GPU stream the kernel was launched on, as modern GPUs can execute multiple streams in parallel~\cite{li2014performance}.
    \item  \textbf{Event Id:} Each CPU or GPU event has unique id.
\end{itemize}

\subsection{Trace Anti-Patterns}\label{sec:anti_pattern}
\begin{figure*}[t]
    \centering
    \includegraphics[width=\linewidth]{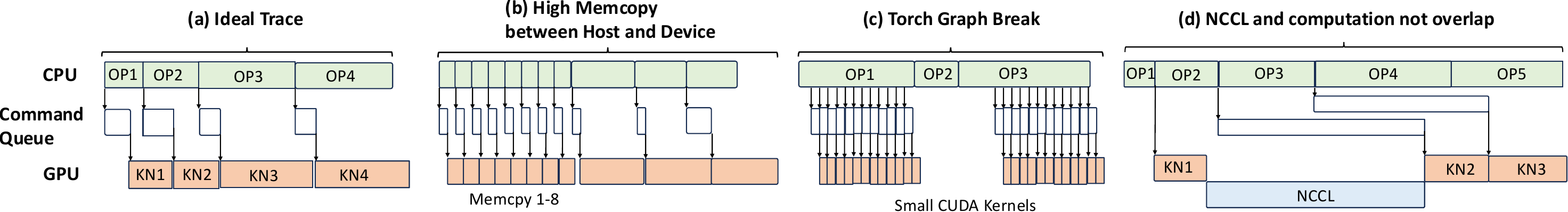}
    \caption{Torch trace anti-patterns illustration. Here OP represents torch CPU operations and KN represents CUDA kernels.}
    \label{fig:anti_patterns}
\end{figure*}

To improve the throughput of ML model training and inference, ML infrastructure engineers often analyze torch traces in a chronological, top-down manner through the call stack. Their goal is to identify inefficient GPU kernels and CPU functions—combinations that we refer to as \textbf{torch trace anti-patterns}. Figure~\ref{fig:anti_patterns} illustrates several common anti-patterns, such as:
\begin{itemize}
    \item Frequent memory copies between CPU and GPU.
    \item PyTorch compile graph break.
    \item Distributed training communication (e.g., NCCL kernels) not overlapping with the compute stream.
\end{itemize}
Traditionally, detecting anti-patterns in torch traces is time-consuming, requiring engineers to manually review thousands of events and apply deep system knowledge to identify and classify suspicious patterns~\cite{bachkaniwala2024lotus}. To address these challenges, we introduce \textbf{TorchTraceAP} in the next section.


\section{Trace Dataset}\label{sec:trace_dataset}
\begin{figure*}[t]
    \centering
    \includegraphics[width=0.9\linewidth]{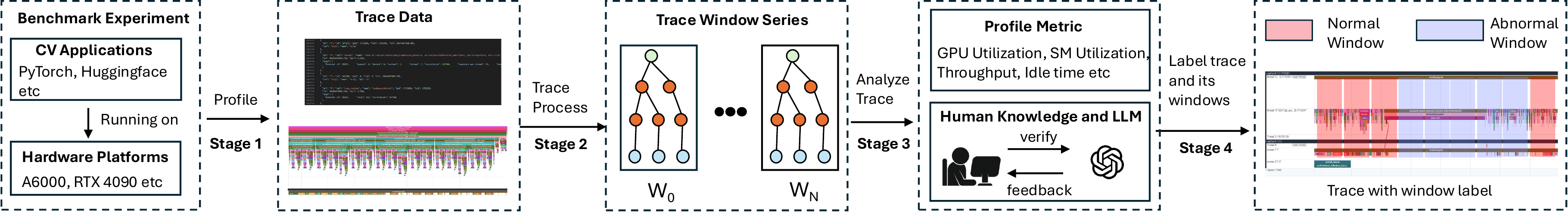}
    \caption{Overview of the Torch Trace Anti-Pattern Benchmark Dataset Collection Pipeline, illustrating the flow from diverse application domains and hardware platforms through trace profiling, window labeling, and human feedback stages to generate a comprehensive benchmark dataset.}
    \label{fig:data_collection}
\end{figure*}

\subsection{Torch-trace Collection and Annotation}

Figure~\ref{fig:data_collection} illustrates the dataset collection pipeline, which now consists of five main stages:

\textbf{Stage 1: Task, Model, and Hardware Platform Selection.} We begin by identifying relevant computer vision applications (downstream tasks) and selecting corresponding model implementations. For example, in object detection, we include both CNN-based methods such as YOLO~\cite{redmon2016you} as well as transformer-based methods like DETR~\cite{carion2020end}. These models are executed on a range of GPU platforms widely used in academia and industry, such as NVIDIA RTX 4090 and A6000. Both single-GPU and multi-GPU training scenarios are included, with the number of GPUs varied from 2 to 8 to capture distributed training anti-patterns.

\textbf{Stage 2: Trace Collection and Preprocessing.} We instrument training and testing loops with the torch profiler to collect traces for 2–3 profiling steps, capturing data loading, forward and backward passes, and loss/gradient computation. As detailed in Section~\ref{sec:trace_background}, raw traces contain numerous CPU and GPU events, each as a JSON entry in PyTorch Kineto, making them difficult for ML models to process. To address this, we segment traces into temporal windows and represent CUDA kernels and their parent CPU kernels as tree structures. This preprocessing allows ML models to encode window information using methods inspired by vision transformers~\cite{dosovitskiy2020image} and GCNs~\cite{kipf2016semi}.

\textbf{Stage 3: Trace Analysis.} We analyze the collected traces using a combination of human expertise, large language models (LLMs), and profiled metrics such as launch latency, SM utilization, throughput, and idle time. Human knowledge and LLMs (e.g., GPT-4.1) equipped with retrieval-augmented generation (RAG) and agent capabilities are leveraged to interpret trace patterns and provide feedback. These tools can reference online discussions and documentation to assist in identifying abnormal behaviors and performance bottlenecks.

\textbf{Stage 4: Trace Labeling.} We label the torch traces from two perspectives. First, we assign a global label to each trace as either a ``good trace'' or a ``bad trace.'' Unlike video anomaly detection, we define a ``good trace'' as one with high throughput and high GPU utilization, and a ``bad trace'' as one with low throughput and low GPU utilization. Profiling throughput and GPU utilization are obtained using PyTorch Holistic Trace Analysis~\cite{torch_hta}. Second, we perform window-level labeling by examining each trace window and cross-validating with LLMs and human feedback.

\subsection{Dataset Metric}

TorchTraceAP dataset comprises two tasks: (1) detecting trace windows containing CPU/GPU anti-patterns, and (2) classifying the specific anti-pattern in each window. For detection, we use the window-level \textit{Area Under the ROC Curve} ($\mathrm{AUC}_{\mathrm{ROC}}$) to measure the model's ability to distinguish windows with anti-patterns. For classification, we report \textit{accuracy}, i.e., the proportion of correctly identified anti-pattern types. Given the scale of candidate kernels, efficient window selection is crucial. This work focuses on the first task, proposing lightweight, weakly-supervised models for window detection. For anti-pattern classification, we utilize existing LLMs (e.g., GPT-4.1).

\section{Methodology} \label{sec:method}
\begin{figure*}
    \centering
    \includegraphics[width=\linewidth]{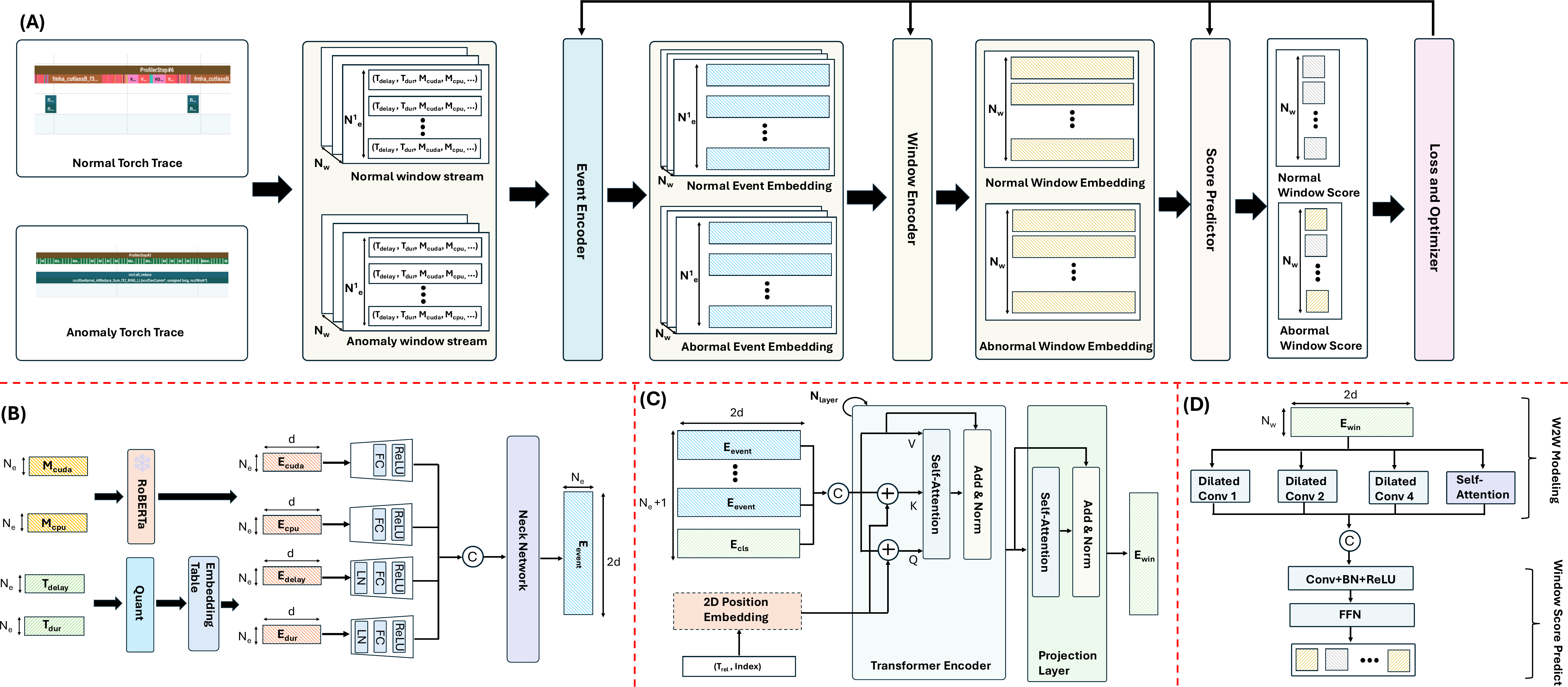}
    \caption{(A) Torch Trace Anti-pattern detection model training flow. (B) Event encoder model architecture. (C) Window encoder model architecture. (D) Window anomaly score predictor model architecture. }
    \label{fig:design}
\end{figure*}

\subsection{Overview and Challenge}



To our knowledge, no existing machine learning methods are tailored for detecting anti-patterns in Torch trace segments. Unsupervised clustering~\cite{campos2016evaluation} (e.g., K-Means) and direct use of LLMs (e.g., GPT-4o) have key limitations: they lack encoders for the hierarchical structure of CPU/GPU events, and LLMs are inefficient with long traces.

To address this, we propose a hierarchical Torch trace encoder inspired by ViT~\cite{dosovitskiy2020image} and GCN~\cite{kipf2016semi}. Our encoder embeds events within time-frame windows and aggregates them into window embeddings. An anomaly detector assigns scores to each window, and anomalous windows are further analyzed by prompting an LLM to classify anti-patterns, enabling iterative, feedback-driven detection.

\subsection{Event Attributes Encode}
Figure~\ref{fig:design}(A) illustrates the overall training flow. Following the previous MIL approach~\cite{sultani2018real}, each training step receives one positive torch trace and one negative torch trace. The negative torch trace contains clear anti-patterns, and exhibits higher latency compared to the positive trace. 

Suppose each trace is divided into $\mathbf{N}_\mathbf{w}$ windows, and for the $i^{th}$ window, there are $\mathbf{N}^{\mathbf{i}}_{\mathbf{e}}$ CUDA kernel events. We select the CUDA kernel event as the anchor point, since each CUDA kernel event is launched by at least one CPU event (as shown in Figure~\ref{fig:background}(B)). Given that the original torch trace event is in plain text and contains multiple types of information, we extract these features into an embedding vector ($\mathbf{E}_\mathbf{event}$). Figure~\ref{fig:design}(B) presents our customized \textbf{event encoder}.

Since anti-patterns arise from the relationships and timing between different events, as shown in Figure~\ref{fig:design}(B), each event encoder fuses the GPU event name (string), CPU event name (string), CPU-to-GPU launch delay (integer), and GPU event duration (integer) into a single embedding ($\mathbf{E}_\mathbf{event}$). For the text-based inputs, after tokenization, we obtain the GPU event ($\mathbf{M}_\mathbf{cuda}$) and CPU event ($\mathbf{M}_\mathbf{cpu}$). A training-free text encoder ($\mathbf{Model}_\mathbf{text}$), such as RoBERTa~\cite{liu2019roberta}, first encodes the tokens into an embedding matrix, which is then passed through a trainable adapter to learn feature representations:
\begin{equation}
\mathbf{E}_\mathbf{cuda} = \mathbf{ReLU}(\mathbf{FC}(\mathbf{Model}_\mathbf{text}(\mathbf{M}_\mathbf{cuda})))
\end{equation}
\begin{equation}
\mathbf{E}_\mathbf{cpu} = \mathbf{ReLU}(\mathbf{FC}(\mathbf{Model}_\mathbf{text}(\mathbf{M}_\mathbf{cpu})))
\end{equation}

Here, \textbf{FC} denotes fully connected layers, and ReLU is the Rectified Linear Unit activation~\cite{gao2024clip}. Both embedding matrices, $\mathbf{E}_\mathbf{cuda}$ and $\mathbf{E}_\mathbf{cpu}$, have shape $N_\text{event} \times d$, where $d$ is the embedding dimension.

For the launch delay and duration of the GPU event ($\mathbf{T}_\mathbf{delay}$ and $\mathbf{T}_\mathbf{dur}$), since their values can be very large, we first apply a log-based quantization method to scale them down. The quantized values are then fed into a trainable embedding table ($\mathbf{Model}_\mathbf{table}$) to generate embedding vectors. These vectors are further processed by an adapter with layer normalization (LN), fully connected layers, and ReLU.
The quantized time \textbf{T'} is computed as:
\begin{equation}
\resizebox{0.43\textwidth}{!}{$
T' = \left\lfloor
    \frac{
        \log\left(\operatorname{clamp}(T, T^{\min}, T^{\max}) + 1\right)
        - \log(T^{\min} + 1)
    }{
        \log(T^{\max} + 1) - \log(T^{\min} + 1)
    }
    \times (N - 1)
\right\rfloor
$}
\end{equation}
where $T^{\min}$ and $T^{\max}$ are the minimum and maximum values of the launch delay and duration, and $N$ is the vocabulary size of the embedding table, and T is the original delay value.

After obtaining the embedding vectors for each attribute, we concatenate them along the channel dimension, resulting in an $N_e \times 4d$ embedding matrix. This matrix is then fed into a convolution-based neck network ($\mathbf{M}_\mathbf{neck}$)~\cite{kirillov2023segment} to extract the final event embedding ($\mathbf{E}_\mathbf{event}$) with shape $N_e \times 2d$:
\begin{equation} \label{eq:neck_net}
\resizebox{0.43\textwidth}{!}{$
\mathbf{E}_\mathbf{event} = \mathbf{Model}_\mathbf{neck}(\mathbf{E}_\mathbf{cuda} \oplus \mathbf{E}_\mathbf{cpu} \oplus \mathbf{E}_\mathbf{delay} \oplus \mathbf{E}_\mathbf{dur})
$}
\end{equation}

\subsection{Window Events Encode}
Figure~\ref{fig:design}(C) presents the architecture of the window encoder model, which fuses event embedding information into a window embedding to represent inter-window event relations. The window encoder consists of three main components: two-dimensional positional encoding, a transformer encoder, and a projection layer.

Similar to the Vision Transformer and text transformer models, the window encoder also includes a trainable [CLS] token, which is concatenated with the event embeddings as follows:
\begin{equation}
    \mathbf{E'}_{\mathbf{event}} = [\text{CLS}] \oplus \mathbf{E}_{\mathbf{event}}
\end{equation}
For positional encoding, we consider that two coordinates are necessary to represent each event's relative position within a window: the relative launch time ($T_{\text{rel}}$) and the stream ID ($I_{\text{stream}}$). The relative launch time is computed as:
\begin{equation}
    T_{\text{rel}} = \min(0, T^{e}_{\text{start}} - T^{w}_{\text{win}})
\end{equation}
Here, $T^{e}_{\text{start}}$ denotes the absolute start time of the event, and $T^{w}_{\text{win}}$ is the window start time. The stream ID ($I_{\text{stream}}$) indicates that, at any given time, several CUDA kernels may execute simultaneously on different streams~\cite{li2014performance}. For example, computation-related GPU CUDA kernels such as cuDNN and CUTLASS may launch on stream 7, while communication-related kernels such as NCCL AllReduce may launch on stream 14. The two-dimensional positional encoding is implemented following RoPE~\cite{su2024roformer}:
\begin{equation}
\resizebox{0.43\textwidth}{!}{
\(
PE(T_{\text{rel}}, I_{\text{stream}}) = 
\left[ 
    \operatorname{SinCos}(T_{\text{rel}} \cdot \text{inv\_freq}),\ 
    \operatorname{SinCos}(I_{\text{stream}} \cdot \text{inv\_freq}) 
\right]_{[:D]}
\)
}
\end{equation}
Here, $\operatorname{SinCos}$ denotes the concatenation of $\sin()$ and $\cos()$ for all elements of the event embedding matrix ($\mathbf{E'}_{\mathbf{event}}$). $D$ is the embedding dimension, which is $2d$ as defined in Equation~\ref{eq:neck_net}. Position embedding is then added to event embedding before feeded into transformer encoder:
\begin{equation}
    \mathbf{E'}_{\mathbf{event}} = \mathbf{E'}_{\mathbf{event}} + \mathbf{PE}
\end{equation}

Next, we feed the event embeddings into a transformer encoder, where each layer consists of a multi-head self-attention (MHSA) mechanism and a feed-forward network (FFN)~\cite{dosovitskiy2020image}.
After the transformer encoder, the event embeddings are passed through a final projection layer, which includes a cross-attention mechanism between the event embeddings and a trainable hardware platform embedding. Finally, we extract the window embedding vector ($\vec{E}_{\text{win}}$).

\begin{figure*}
    \centering
    \includegraphics[width=\linewidth]{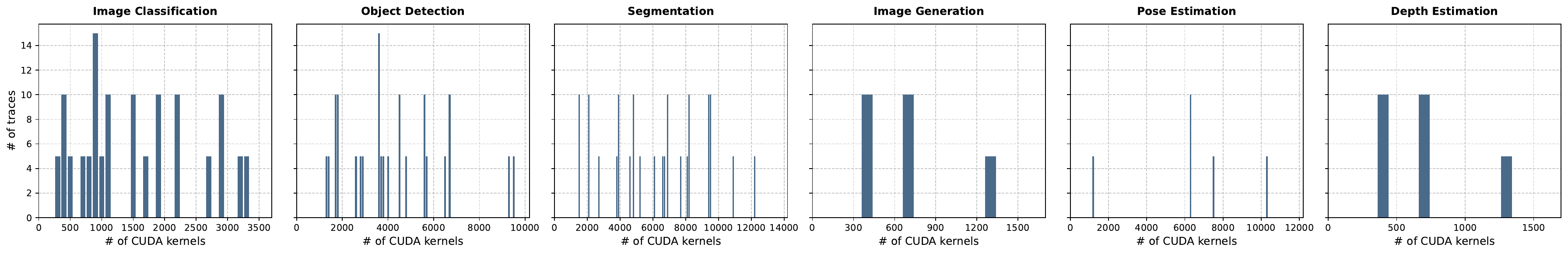}
    \caption{TorchTraceAP applications CUDA kernel number distribution.}
    \label{fig:trace_dist}
\end{figure*}


\subsection{Window-to-Window Detection}
We concatenate window embeddings and apply 1D positional encoding~\cite{dosovitskiy2020image} to form the matrix $\mathbf{E}_{\text{event}}$ ($N_{\text{win}} \times D$). To identify windows with potential anti-patterns, we capture local relationships using dilated convolutions (DiConv)~\cite{yu2015multi} and global relationships with self-attention~\cite{vaswani2017attention}:
\begin{equation}
\resizebox{0.43\textwidth}{!}{
    $\mathbf{E'}_{\text{win}} = \text{DiConv}(E_{\text{win}}, 1) \oplus \text{DiConv}(E_{\text{win}}, 2) \oplus \text{DiConv}(E_{\text{win}}, 4) \oplus \text{MHSA}(E_{\text{win}})$
}
\end{equation}
Here, $\text{DiConv}(\mathbf{X}, i)$ is a 1D dilated convolution with dilation $i$. We then use convolutional layers to reduce the dimension from $4d$ to $2d$, and a feed-forward network (FFN) to produce anomaly scores $\mathbf{P}_{\text{w}}$ ($1 \times N_{\text{win}}$). Windows with scores above a threshold are flagged as containing anti-patterns and sent to an LLM for further analysis.

\subsection{End to End Training Loss}
For \textbf{weakly supervised training}, we provide only a trace-level label $\mathbf{Y}$ for each trace, where $\mathbf{Y}^{+}$ denotes a positive trace and $\mathbf{Y}^{-}$ denotes a negative trace. Our training framework follows previous MIL-based methods~\cite{sultani2018real, tian2021weakly}, with several key modifications. Specifically, rather than assuming that normal traces contain no anti-patterns, we allow for the possibility that positive traces may still include a limited number of anti-patterns. Consequently, instead of using the highest window anomaly score, we use the average score across the entire trace as the trace-level score, and define the trace-level loss ($\mathbf{L}_{\text{trace}}$) as follows:
\begin{equation}
    \mathbf{L}_{\text{trace}} = \text{BCELoss}\left(\text{avg}(\mathbf{P}_{\text{w}}),\ \mathbf{Y}\right)
\end{equation}
Here, \textbf{BCELoss} denotes the binary cross-entropy loss. To further separate the embedding spaces of anomalous and normal windows, we select the top $S_{+}\%$ of normal trace window embedding vectors and the bottom $S_{-}\%$ of abnormal trace window embedding vectors, and apply an SVM-inspired loss ($\mathbf{L}_{\text{svm}}$):
\begin{equation}
\resizebox{0.43\textwidth}{!}{$
    \mathbf{L}_{\text{svm}} = \frac{1}{N_{+} N_{-}} \sum_{i=1}^{N_{+}} \sum_{j=1}^{N_{-}} \max\left(0,\ \mathrm{margin} - \left( \frac{\mathbf{n}_i}{\|\mathbf{n}_i\|_2} \cdot \frac{\mathbf{a}_j}{\|\mathbf{a}_j\|_2} \right) \right)
$}
\end{equation}
\begin{equation}
    \mathbf{n}_i = \text{top}(S_{+},\ \mathbf{E}'_{\text{win+}})[i],\quad i \in [0, N_{+})
\end{equation}
\begin{equation}
    \mathbf{a}_i = \text{top}(S_{-},\ \mathbf{E}'_{\text{win-}})[i],\quad i \in [0, N_{-})
\end{equation}
where $N_{+} = S_{+} \times N_{\text{win}}$ and $N_{-} = S_{-} \times N_{\text{win}}$. For the remaining $N_{\text{win}} - N_{+}$ and $N_{\text{win}} - N_{-}$ window embeddings, we apply a clustering-based loss ($\mathbf{L}_{\text{cluster}}$) to encourage embeddings with the same attribute (presence or absence of anti-patterns) to be close, and those with different attributes to be separated:
\begin{equation}
\resizebox{0.43\textwidth}{!}{$
\mathcal{L}_{\text{cluster}} = \frac{1}{|P|} \sum_{(i,j) \in P} \max(0, m - S_{ij}) + \frac{1}{|N|} \sum_{(i,j) \in N} \max(0, S_{ij} + m) + \frac{1}{M} \sum_{i \neq j} \left| S_{ij} - D_{ij} \right|
$}
\end{equation}

Here, $S_{ij}$ denotes the similarity between window embeddings $i$ and $j$, and $D_{ij}$ is the pairwise window label distance. Since our approach is weakly supervised, we do not use actual window-level labels; instead, we use the predictor's score ($\mathbf{P}_{\text{w}}$) as a pseudo-window label. 

We also apply a smoothness loss ($\mathbf{L}_{\text{smooth}}$), following prior work in time-series and video anomaly detection~\cite{lee2021weakly, sultani2018real}, to reduce abrupt feature changes between consecutive windows. Combining all components, the final training loss ($\mathbf{L}$) is defined as:
\begin{equation}
\resizebox{0.43\textwidth}{!}{$
    \mathbf{L} = \alpha_1 \cdot \mathbf{L}_{\text{trace}} + \alpha_2 \cdot \mathbf{L}_{\text{svm}} + \alpha_3 \cdot \mathbf{L}_{\text{cluster}} + \alpha_4 \cdot \mathbf{L}_{\text{smooth}}
$}
\end{equation}
\subsection{LLM-based Fine-grained Classification}
After detecting suspicious Torch trace windows, we provide the event information from these windows to a large language model (LLM). The specific prompt used for the LLM is included in the Appendix. This process serves two main purposes: (1) to identify any anti-patterns present within the trace window, and (2) to assess whether the window may have been misclassified as anomalous. If misclassification is suspected, we consider adjusting the window size and conducting a second round of detection. This approach leverages the LLM’s ability to interpret complex event data, thereby refining our anomaly detection pipeline and improving overall accuracy. The prompt provided to the LLM is detailed in the Appendix.

\section{Experiments}
\subsection{Dataset Attributes}
\begin{table}[]
\caption{TorchTraceAP vision applications and models.}
\label{tab:vision_model_sum}
\resizebox{0.48\textwidth}{!}{%
\begin{tabular}{c|c}
\toprule
                     & \textbf{Models}                                                                                    \\ \midrule
Image Classification & VGG~\cite{simonyan2014very}, ResNet~\cite{he2016deep}, ViT~\cite{dosovitskiy2020image}, Swin~\cite{liu2021swin}                                                                           \\ \midrule
Object Detection     & \begin{tabular}[c]{@{}c@{}}Yolo~\cite{redmon2016you}, Detectron2~\cite{wu2019detectron2}, DETR~\cite{carion2020end}, \\ Deformable\_DETR~\cite{zhu2020deformable}, ViTDet~\cite{li2022exploring}, YOLOS~\cite{fang2021you}\end{tabular} \\ \midrule
Image Segmentation   & \begin{tabular}[c]{@{}c@{}} Segformer~\cite{xie2021segformer}, Yolo~\cite{redmon2016you}, Detecron2~\cite{wu2019detectron2}, \\  mask2former~\cite{cheng2022masked}, SAM~\cite{kirillov2023segment}\end{tabular}                                                       \\ \midrule
Image Generation     & Stable Diffusion~\cite{rombach2022high}, GAN~\cite{goodfellow2020generative}, DDPM~\cite{ho2020denoising},                                                                       \\ \midrule
Depth Estimation     & Depth Anything V2~\cite{yang2024depth}, DPT~\cite{ranftl2021vision},  ZoeDepth~\cite{bhat2023zoedepth}                                                                  \\ \midrule
Pose Estimation      & Yolo~\cite{redmon2016you}                                                                                               \\ \bottomrule
\end{tabular}%
}
\end{table}

\begin{table}[]
\centering
\caption{TorchTraceAP anti-pattern types distribution on testing split. Please check Appendix for each anti-pattern meaning and visualization. }
\label{tab:anti_pattern_dist}
\resizebox{0.4\textwidth}{!}{%
\begin{tabular}{ccc}
\toprule
Index & Anti-pattern Type            & $N_{event}$ \\ \midrule
1     & Single Process Dataloading   & 221       \\
2     & No Memory Pin                & 193       \\
3     & High Memcpy of H2D and D2H   & 334       \\
4     & Torch Graph Break            & 74        \\
5     & CPU Bound                    & 271       \\
6     & Samll NCCL Kernels           & 312       \\
7     & NCCL Block main Stream       & 65        \\
8     & NCCL and Compute not Overlap & 103       \\ \bottomrule
\end{tabular}%
}
\end{table}

TorchTraceAP dataset encompasses 6 distinct common vision-related applications, including image classification, object detection, image segmentation, image generation, depth estimation, and pose estimation. We profile these models on various GPU platforms including A6000, RTX 4090, A6000 Ada, A100, and H100 with both single GPU and multiple GPU settings.  Some models are further optimized using acceleration frameworks (such as DeepSpeed~\cite{rasley2020deepspeed} and Huggingface Accelerate~\cite{wolf2020transformers}). \textbf{All the Torch trace that we collect are from publicly available models or frameworks.} We present each applications and models details in Table~\ref{tab:vision_model_sum}.

For the training set, we provide torch traces for image classification, object detection, and image segmentation tasks. The test set is divided into seen-test and unseen-test categories. The seen-test set includes torch traces from the same tasks as training (image classification, object detection, and image segmentation). In contrast, the unseen-test set contains torch traces from image generation, pose estimation, and action recognition tasks. Compared to the training traces, the unseen-test traces feature models with similar backbones (such as ResNet~\cite{he2016deep} or ViT~\cite{dosovitskiy2020image}) but different detection heads, decoder architectures, or specialized modules (e.g., temporal memory modules for action recognition). In total, the dataset contains 610 torch traces, comprising 460 training traces and 150 testing traces. For the training set, only trace-level labels are provided, while for the testing set, both trace-level and window-level labels are available. Figure~\ref{fig:trace_dist} summarizes the distribution of torch traces across vision applications for both training and testing. Table~\ref{tab:anti_pattern_dist} summarizes for testing trace, different anti-pattern types distribution. We present complete visualizations of each anti-pattern in the Appendix.

\begin{table*}[]
\centering
\caption{Window-level anti-pattern detection results. We use AUC-ROC (\%) as the evaluation metric. The test data is divided into `seen' and `unseen' groups. Various numbers of windows ($N_w$) are tested.}
\label{tab:det_res}
\resizebox{0.95\textwidth}{!}{%
\begin{tabular}{ccccccccc}
\toprule
\multicolumn{1}{l|}{}              & \multicolumn{4}{c|}{\textbf{Seen}}                                                                                             & \multicolumn{4}{c}{\textbf{Unseen}}                                                               \\ \midrule
\multicolumn{1}{l|}{}              & \textbf{Image Classification} & \textbf{Object Detectuion} & \textbf{Image Segmentation} & \multicolumn{1}{c|}{\textbf{Avg}}   & \textbf{Pose Estimation} & \textbf{Depth Estimation} & \textbf{Image Generation} & \textbf{Avg}   \\ \midrule
\multicolumn{1}{l}{}               & \multicolumn{8}{c}{\textbf{$N_{w}$ = 100}}                                                                                                                                                                                          \\ \midrule
\multicolumn{1}{c|}{Rule}          & 52.13                         & 51.34                      & 51.68                       & \multicolumn{1}{c|}{51.71}          & 54.19                    & 53.12                     & 53.11                     & 52.46          \\
\multicolumn{1}{c|}{Clustering}    & 50.86                         & 50.91                      & 50.19                       & \multicolumn{1}{c|}{50.65}          & 50.21                    & 50.48                     & 50.51                     & 50.54          \\
\multicolumn{1}{c|}{Pure LLM}      & 56.11                         & 54.32                      & 52.69                       & \multicolumn{1}{c|}{54.37}          & 55.31                    & 54.43                     & 53.32                     & 54.36          \\
\multicolumn{1}{c|}{\textbf{Ours}} & \textbf{84.28}                & \textbf{76.79}             & \textbf{74.15}              & \multicolumn{1}{c|}{\textbf{78.41}} & \textbf{65.22}           & \textbf{61.47}            & \textbf{63.57}            & \textbf{63.42} \\ \midrule
\multicolumn{1}{l}{}               & \multicolumn{8}{c}{\textbf{$N_{w}$ = 50}}                                                                                                                                                                                           \\ \midrule
\multicolumn{1}{c|}{Rule}          & 51.27                         & 52.45                      & 51.93                       & \multicolumn{1}{c|}{51.88}          & 52.47                    & 53.92                     & 54.31                     & 53.56          \\
\multicolumn{1}{c|}{Clustering}    & 51.11                         & 50.21                      & 50.49                       & \multicolumn{1}{c|}{50.60}          & 50.93                    & 50.21                     & 51.01                     & 50.71          \\
\multicolumn{1}{c|}{Pure LLM}      & 53.39                         & 52.81                      & 54.12                       & \multicolumn{1}{c|}{53.44}          & 53.76                    & 54.29                     & 56.82                     & 54.95          \\
\multicolumn{1}{c|}{\textbf{Ours}} & \textbf{80.52}                & \textbf{69.53}             & \textbf{69.26}              & \multicolumn{1}{c|}{\textbf{73.10}} & \textbf{63.18}           & \textbf{63.12}            & \textbf{67.12}            & \textbf{64.47} \\ \midrule
\multicolumn{1}{l}{}               & \multicolumn{8}{c}{\textbf{$N_{w}$ = 200}}                                                                                                                                                                                          \\ \midrule
\multicolumn{1}{c|}{Rule}          & 51.73                         & 52.03                      & 51.57                       & \multicolumn{1}{c|}{51.77}          & 52.38                    & 53.33                     & 52.16                     & 52.62          \\
\multicolumn{1}{c|}{Clustering}    & 50.67                         & 51.12                      & 50.83                       & \multicolumn{1}{c|}{50.87}          & 50.63                    & 51.02                     & 50.74                     & 50.79          \\
\multicolumn{1}{c|}{Pure LLM}      & 54.26                         & 54.15                      & 55.19                       & \multicolumn{1}{c|}{54.53}          & 53.31                    & 55.45                     & 53.79                     & 54.18          \\
\multicolumn{1}{c|}{\textbf{Ours}} & \textbf{74.24}                & \textbf{68.85}             & \textbf{81.85}              & \multicolumn{1}{c|}{\textbf{74.98}} & \textbf{60.64}                    & \textbf{63.09}                     & \textbf{63.07}                     & \textbf{62.26}          \\ \bottomrule
\end{tabular}%
}
\end{table*}

\begin{figure*}
    \centering
    \includegraphics[width=0.9\linewidth]{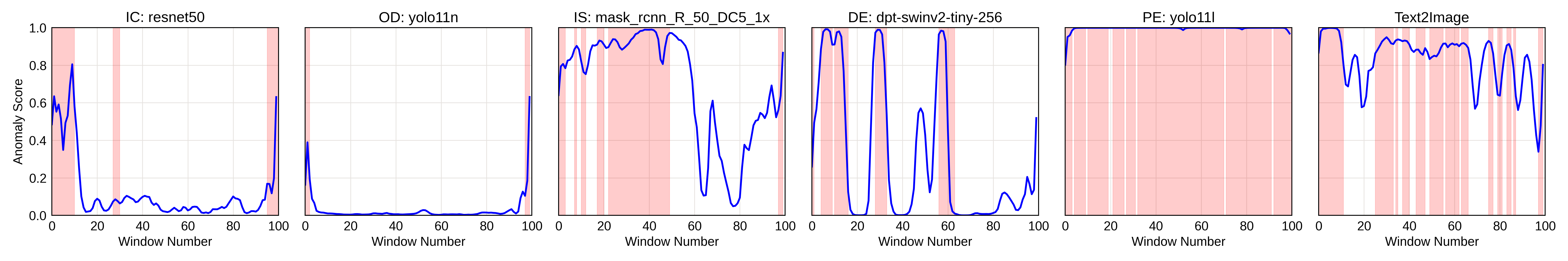}
    \caption{Anomaly scores of our method on different torch trace. Pink areas indicate the manually labelled trace windows with anti-pattern. Here IC represents image classification, OD represents object detection, IS represents image segmentation, PE represents pose estimation, DE represents depth estimation, and IG represents image generation.}
    \label{fig:visualize}
\end{figure*}

\subsection{Experiment Setup}
For the framework proposed in Section~\ref{sec:method}, we use RoBERTa~\cite{liu2019roberta} as a training-free text encoder. The embedding dimension for each event attribute (\textbf{d}) is set to 128, while the window embedding dimension (\textbf{D}) is set to 256. 
For the window encoder, we employ a stack of 6 transformer layers, each equipped with self-attention using 8 heads and a feed-forward network (FFN). The models are optimized using AdamW with a learning rate of 0.0001 and a batch size of 8. 
All experiments are implemented in PyTorch~\cite{paszke2019pytorch} and conducted on a server equipped with a single H100 GPU. For the window anti-pattern fine-grained classification, we choose GPT-4.1 as backend LLM.

\subsection{Anti-pattern Detection Result}

\begin{table}[]
\centering
\caption{LLM fine-grained torch anti-pattern classification result.}
\label{tab:finegrain_classification}
\resizebox{0.4\textwidth}{!}{%
\begin{tabular}{c|ccc}
\toprule
\textbf{Application}          & \multicolumn{3}{c}{\textbf{Number of Windows ($N_{win}$)}} \\ \midrule
\multicolumn{1}{l|}{}         & 50         & 100        & 200       \\ \midrule
Image Classification & 79.17      & 80.12      & 79.29     \\
Object Detectuion    & 65.53      & 66.52      & 64.31     \\
Image Segmentation   & 73.96      & 71.75      & 71.38     \\
Pose Estimation      & 68.52      & 67.63      & 67.25     \\
Depth Estimation     & 80.19      & 81.8       & 81.3      \\
Image Generation    & 86.5       & 84.5       & 85.9      \\ \midrule
Avg                  & 75.64      & 78.9       & 74.91     \\ \bottomrule
\end{tabular}%
}
\end{table}

Table~\ref{tab:det_res} presents the performance of various approaches on TorchTraeAP with different numbers of window splits. For the baseline, we select three methods: a rule-based approach (\textbf{Rule}), an unsupervised clustering-based approach (\textbf{Clustering}), and a pure LLM-based approach (\textbf{LLM}). The Rule method utilizes PyTorch HTA~\cite{torch_hta} to model the distribution of torch CUDA event runtimes, identifying outlier events as abnormal. The Clustering method applies K-means clustering to encoded window embeddings and similarly treats outlier events as abnormal. For the LLM method, we use GPT-4.1 as the baseline. Due to the context window limitations of LLMs, each trace is split into five sub-traces, which are fed directly into GPT-4.1 to obtain an anomaly score for each window. A prompt screenshot for the LLM is provided in the Appendix.

As shown in Table~\ref{tab:det_res}, the baseline methods, including the LLM-based approach, do not perform as well as our proposed method. The rule-based approach cannot generalize to different models within the same vision task; for example, the CUDA event distribution varies when switching from a ResNet-based model to a transformer-based model. The LLM-based method strictly follows specific rules, such as enforcing a threshold for launch delay between CPU and CUDA events. Without prompt-tuning, the LLM also relies on these static rules to detect anti-patterns. Without contrastive training using both positive and negative traces, these static models struggle to generalize across diverse computer vision applications and anti-patterns.

In contrast, our method achieves higher accuracy by applying a weakly supervised approach to model the relationships between different events at multiple levels. We also observe that, for more complex models and tasks (such as image generation and segmentation), increasing $N_w$ improves detection accuracy. After identifying torch trace windows with anti-patterns, we prompt GPT-4.1 to perform fine-grained classification of the specific anti-pattern type. As shown in Table~\ref{tab:finegrain_classification}, when directly prompted to classify the anti-pattern type of a trace window, the LLM achieves high reasoning accuracy. We believe that explicitly prompting the LLM with the information that a window contains an anti-pattern provides clearer reasoning direction.

\subsection{Ablation study and Visualization}
\begin{table}[]
\centering
\caption{Ablation study for different module.}
\label{tab:ablation_module}
\resizebox{0.45\textwidth}{!}{%
\begin{tabular}{ccc|cc}
\toprule
Event Adapter                        & Transformer Encoder                        & Relation Modelting                        & Seen AUC (\%) & Unseen AUC (\%) \\ \midrule
-                         & -                         & -                         & 50.3     & 50.8       \\
\checkmark & -                         & -                         & 50.2     & 50.3       \\
-                         & \checkmark & -                         & 58.26    & 55.81      \\
-                         & -                         & \checkmark & 54.35    & 45.56      \\
\checkmark & \checkmark & -                         & 71.17    & 61.62      \\
\checkmark & -                         & \checkmark & 57.66    & 45.56      \\
-                         & \checkmark & \checkmark & 67.4     & 60.6       \\ \midrule
\checkmark & \checkmark & \checkmark & 78.41    & 63.42      \\ \bottomrule
\end{tabular}%
}
\end{table}

Table~\ref{tab:ablation_module} presents different components effect of our method to the anti-pattern detection accuracy. Our experiment shows that Transformer Encoder at Figure~\ref{fig:design}.(C) and Relation Modeling at Figure~\ref{fig:design}.(D) play pivot role in modeling relation of events inside a window and window to window relations. Without such modeling, the detectin accuracy drop significantly. In Figure~\ref{fig:visualize}, we show the anomaly scores produced by our method on 6 different computer vision applications. For each application, we select one model's trace.


\section{Conclusion}
We present TorchTraceAP, a benchmark and two-stage method for detecting performance anti-patterns in computer vision model torch traces. By combining lightweight ML models with LLM-based analysis, our approach outperforms traditional techniques and makes performance optimization more accessible and automated.

\newpage
{
    \small
    \bibliographystyle{ieeenat_fullname}
    \bibliography{main}
}



\end{document}